\documentclass[11pt, a4paper, logo, copyright, nonumbering]{antgroup}
\usepackage[utf8]{inputenc}
\usepackage[square, numbers]{natbib}
\usepackage{dblfloatfix}
\usepackage{ulem}
\usepackage{caption}
\usepackage{dramatist}
\usepackage{xspace}
\usepackage{pifont}
\usepackage{multirow}

\usepackage{xltabular}
\usepackage{longtable}
\usepackage{hyperref}
\usepackage{amsfonts}
\usepackage{amsmath}
\usepackage{amssymb}
\usepackage{lineno}
\usepackage{adjustbox}

\usepackage[bottom]{footmisc}

\usepackage{CJKutf8}
\usepackage{subfigure}
\usepackage{setspace}

\usepackage{dsfont}
\usepackage{array}
\usepackage{tabularx}
\usepackage{subfigure}
\usepackage{xcolor}
\usepackage{tabularx}
\usepackage{booktabs}
\usepackage{subcaption}
\usepackage{graphicx}

\usepackage{lipsum}
\usepackage{multicol}
\usepackage{authblk}
\usepackage{wrapfig}
\usepackage[most,skins,theorems]{tcolorbox}
\usepackage{array}
\usepackage{pifont}
\definecolor{customgray}{RGB}{230,230,230}

\usepackage{graphicx}
\usepackage{multirow}
\usepackage{booktabs}
\usepackage[table]{xcolor}
\usepackage{hyperref}
\hypersetup{
	colorlinks=true,
	linkcolor=red,   
	urlcolor=blue,
}
\usepackage{array}
\usepackage{amssymb}
\usepackage{makecell}
\usepackage{graphicx}
\usepackage{adjustbox}
\usepackage{pifont}
\usepackage{subcaption}
\usepackage{caption}

\tcbset{
  aibox/.style={
    width=\linewidth,
    top=8pt,
    bottom=4pt,
    colback=gray!10!white,
    colframe=gray!50!black,
    colbacktitle=gray!70!black,
    enhanced,
    center,
    attach boxed title to top left={yshift=-0.1in,xshift=0.15in},
    boxed title style={boxrule=0pt,colframe=white,},
  }
}
\newtcolorbox{AIbox}[2][]{aibox,title=#2,#1}

\makeatletter
\def\@BTrule[#1]{%
  \ifx\longtable\undefined
    \let\@BTswitch\@BTnormal
  \else\ifx\hline\LT@hline
    \nobreak
    \let\@BTswitch\@BLTrule
  \else
     \let\@BTswitch\@BTnormal
  \fi\fi
  \global\@thisrulewidth=#1\relax
  \ifnum\@thisruleclass=\tw@\vskip\@aboverulesep\else
  \ifnum\@lastruleclass=\z@\vskip\@aboverulesep\else
  \ifnum\@lastruleclass=\@ne\vskip\doublerulesep\fi\fi\fi
  \@BTswitch}
\makeatother

\addto\extrasenglish{
}

 {\begin{list}{}%
         {\setlength{\leftmargin}{#1}}%
         \item[]%
 }
 {\end{list}}

\reportnumber{001} %

\renewcommand{\today}{}

\title{\centering VisBrowse-Bench: Benchmarking Visual-Native Search for Multimodal Browsing Agents}

\author{
Zhengbo Zhang$^{1,2, *, \dag}$,
Jinbo Su$^{3,2,*, \dag}$,
Zhaowen Zhou$^{2,*}$,
Changtao Miao$^{2,*}$,
Yuhan Hong$^{4}$,

\vspace{-6pt}
Qimeng Wu$^{5}$,
Yumeng Liu$^{6}$,
Feier Wu$^{5}$,
Yihe Tian$^{5}$,
Yuhao Liang$^{7}$,
Zitong Shan$^{2}$,

\vspace{-6pt}
Wanke Xia$^{5}$,
Yi-Fan Zhang$^{1}$,
Bo Zhang$^{2}$,
Zhe Li$^{2}$,
Shiming Xiang$^{1, \ddag}$,
Ying Yan$^{2, \ddag}$
\\
\vspace{-6pt}
$^{1}$ CASIA, $^{2}$Ant Digital Technologies, Ant Group, $^{3}$ RUC, $^{4}$ FZU, $^{5}$ THU, $^{6}$ USTB, $^{7}$ PKU,
\vspace{-6pt}
}

\footnotetext[1]{Equal contribution.}
\footnotetext[2]{Work done during internship in Ant Group.}
\footnotetext[3]{Corresponding to smxiang@nlpr.ia.ac.cn, fuying.yy@antgroup.com}

\renewcommand{\phi}{\varphi}

\renewcommand{\epsilon}{\varepsilon}
\renewcommand{\imath}{\mathrm{i}}

\newlength{\restsubwidth}
\newlength{\restsubheight}
\newlength{\restsubmoreheight}
\newcommand{\rest}[2]{%
        \settowidth{\restsubwidth}{\ensuremath{#2}}
        \settoheight{\restsubheight}{\ensuremath{{}_{#2}}}
        \ensuremath{{#1\hskip 0.5pt}_{\vrule\kern2pt\parbox[b][%
        4pt][b]{\the\restsubwidth}{%
                        \ensuremath{{}_{#2}}}}}
        }

\begin{abstract}
The rapid advancement of Multimodal Large Language Models (MLLMs) has enabled browsing agents to acquire and reason over multimodal information in the real world. But existing benchmarks suffer from two limitations: insufficient evaluation of visual reasoning ability and the neglect of native visual information of web pages in the reasoning chains. To address these challenges, we introduce a new benchmark for visual-native search, VisBrowse-Bench. It contains 169 VQA instances covering multiple domains and evaluates the models' visual reasoning capabilities during the search process through multimodal evidence cross-validation via text-image retrieval and joint reasoning. These data were constructed by human experts using a multi-stage pipeline and underwent rigorous manual verification.  We additionally propose an agent workflow that can effectively drive the browsing agent to actively collect and reason over visual information during the search process. We comprehensively evaluated both open-source and closed-source models in this workflow.
Experimental results show that even the best-performing model, Claude-4.6-Opus only achieves an accuracy of 47.6\%, while the proprietary Deep Research model, o3-deep-research only achieves an accuracy of 41.1\%. The code and data can be accessed at: \href{https://github.com/ZhengboZhang/VisBrowse-Bench}{https://github.com/ZhengboZhang/VisBrowse-Bench}
\end{abstract}

\begin{document}
\maketitle

\section{Introduction}

\begin{figure*}[t]
    \centering
    \includegraphics[width=1\linewidth]{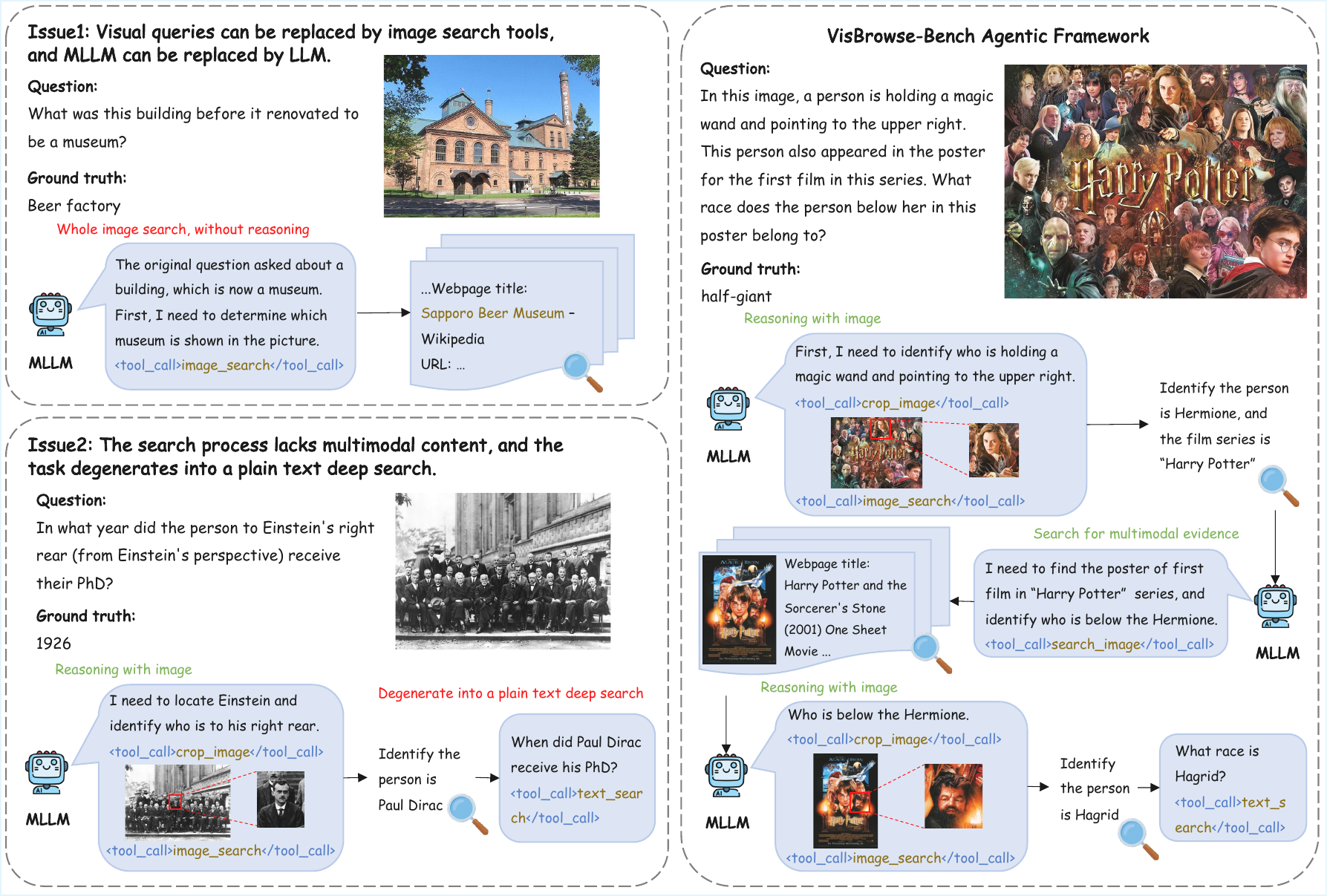}
    \caption{Existing benchmarks have two limitations in evaluating multimodal browsing agents: 1. the semantic information of visual queries can be easily obtained through image search tools; 2. real-world browsing environments contain a wealth of multimodal information, which most benchmarks overlook. VisBrowse-Bench is designed to fuse multimodal information during the search process and ensure that visual capabilities are essential for completing the task.}
    \label{fig1}
\end{figure*}

Driven by the rapid advancements in large language models (LLMs) and agent technologies, a plethora of high-quality works has emerged in the deep research domain~\citep{gunjal2025rubrics,deepeyesv2,shao2025dr,yao2026researcher}. However, existing deep research benchmarks predominantly focus on the textual modality, thereby neglecting the multimodal demands inherent in real-world retrieval scenarios. Concurrently, the evolution of multimodal large language models (MLLMs) has inspired a series of works on Multimodal Browsing Agents~\citep{yao2026mm,yan2025comprehensive}. Nevertheless, existing multimodal benchmarks still exhibit significant limitations, as shown in Figure \ref{fig1}.

Specifically, most current benchmarks (e.g., MMSearch~\citep{mmsearch} and BrowseComp-VL~\citep{webwatcher}) merely test models' ability to invoke tools for solving text-image queries. These tasks typically introduce an image search tool, where models simply feed the image into the tool for retrieval. Such tasks do not require fine-grained understanding of multimodal information and thus fail to sufficiently challenge models' multimodal comprehension capabilities in deep research scenarios, instead primarily emphasizing tool-calling abilities.

Moreover, even though some benchmarks (e.g., MMSearch-Plus ~\citep{mmsearch-plus} and VDR-Bench ~\citep{vdr}) require initial visual perception of text-image queries, the subsequent information gathering process degenerates into single-modal text traversal. Existing benchmarks structure their search space such that once the query image yields an entity name or caption, all downstream reasoning can be completed through textual document retrieval and synthesis. The search trajectory never necessitates grounding, parsing, or reasoning over additional visual information discovered during the search process. The task thus degenerates into text-only browsing, failing to assess whether models can dynamically acquire and integrate visual information when it is not provided upfront but must be actively sought across web pages containing multimodal information. This ability to conduct visual search and understanding throughout the task is vital in real-world retrieval, but existing benchmarks fail to evaluate this capability, limiting further development.

\begin{figure*}[ht]
    \centering
    \begin{minipage}{0.7\textwidth}
        \centering
        \subfigure[]{\includegraphics[width=\linewidth]{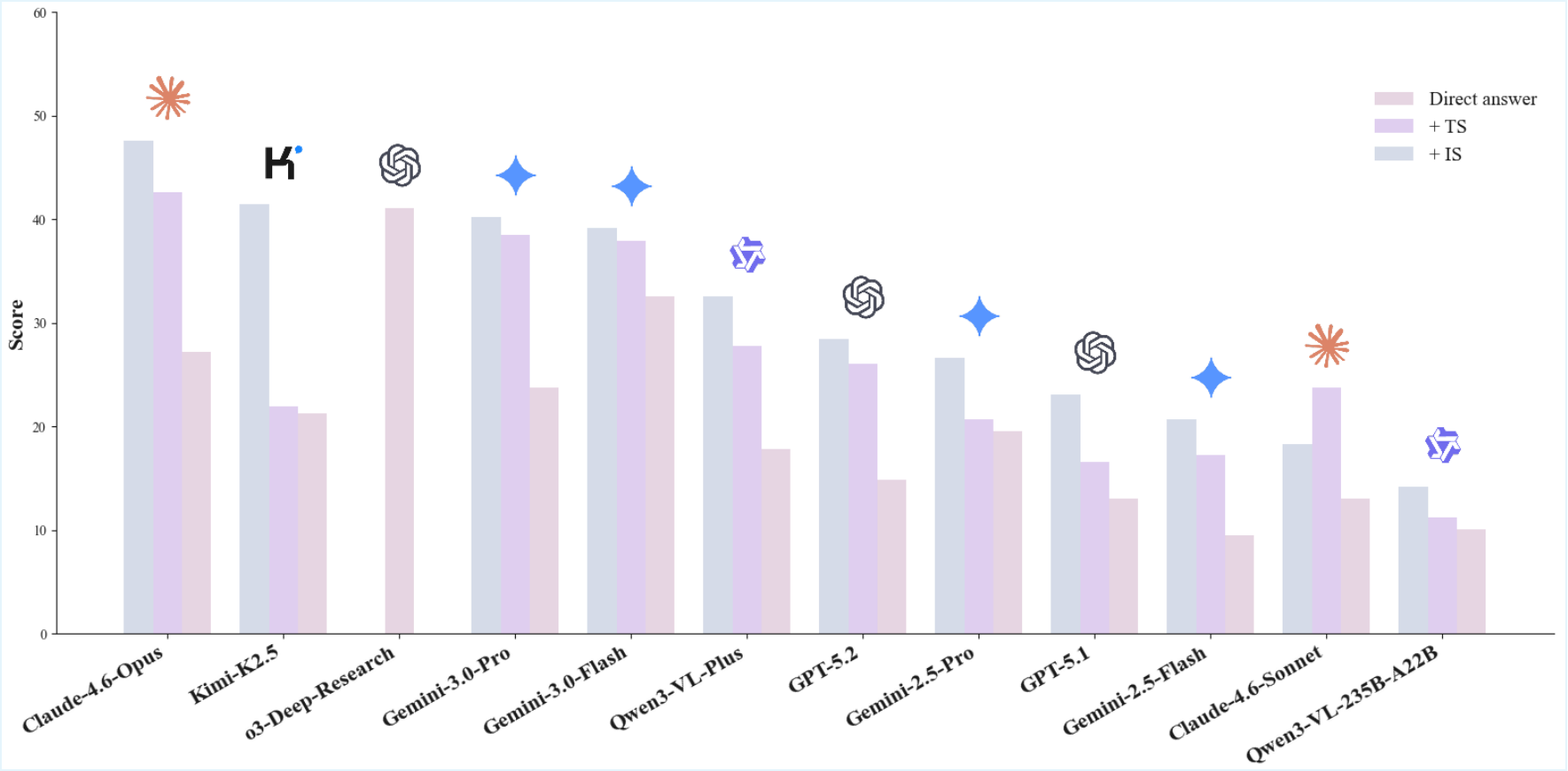}}
    \end{minipage}
    \begin{minipage}{0.28\textwidth}
        \subfigure[]{\includegraphics[width=\linewidth]{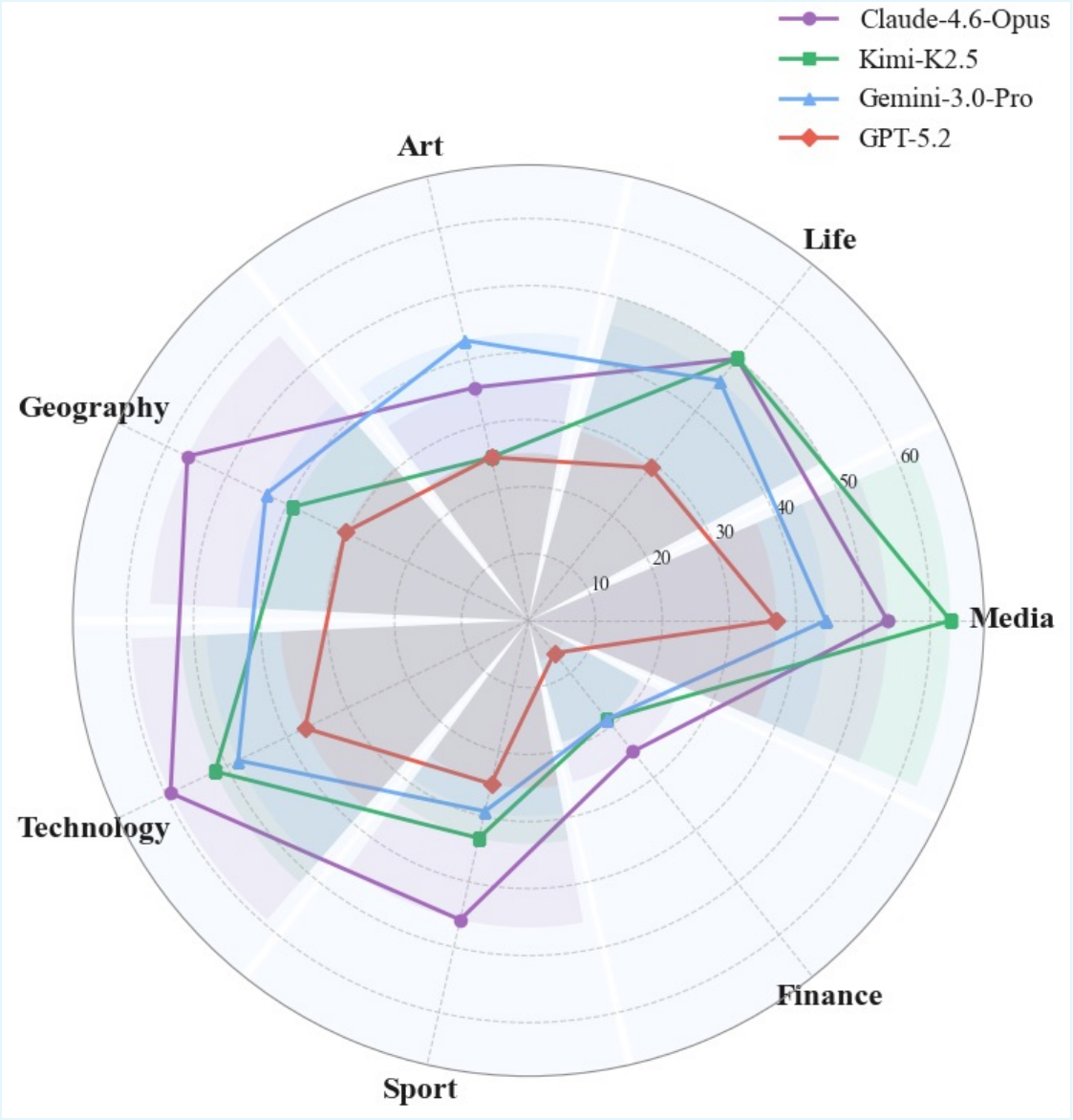}}
    \end{minipage}
    \caption{(a) Overall performance of MLLMs on VisBrowse-Bench. (b) Performance of four MLLMs on seven categories.}
    \label{fig2}
\end{figure*}

To address these challenges, we introduce \textbf{VisBrowse-Bench}, a challenging benchmark to comprehensively evaluate the reasoning and search capability of multimodal browsing agents. VisBrowse-Bench relies on two core design principles: the integration of multimodal information within the reasoning chain, and an inherent dependence on visual capabilities. We constructed our benchmark through a multi-stage, expert-guided pipeline, resulting in 169 VQA instances that cover seven distinct domains. Based on a seed entity with inherent visual ambiguity, domain experts recursively construct multi-hop reasoning chains, which require locating and reasoning novel visual evidence that cannot be paraphrased textually, alongside textual sources that provide complementary but insufficient information. Furthermore, we propose an efficient agentic workflow for visual reasoning and visual information retrieval during the search process. In this workflow, the agent is driven to actively perform visual reasoning and visual information retrieval using a rich set of tools. The preview performance of the MLLMs using our workflow on VisBrowse-Bench is shown in Figure \ref{fig2}. The best performance in VisBrowse-Bench is achieved by Claude-4.6-Opus, with an accuracy of 47.6\%, while most models achieve an accuracy of around 30\%.

Our key contributions can be summarized as follows:

\begin{itemize}
    \item We formalize the task of multimodal browsing, identifying two critical challenges in existing benchmarks: insufficient evaluation of visual reasoning ability and the neglect of visual-native information in the reasoning chains.
    \item We propose VisBrowse-Bench, a new benchmark comprising 169 rigorously validated instances constructed by human experts. This benchmark aims to jointly evaluate the search and visual reasoning capabilities of multimodal browsing agent systems.
    \item We introduce a multi-turn visual information retrieval-reasoning agentic workflow to solve multimodal browsing problems in the real world. Compared to direct answer, the models' performance are significantly improved using our workflow, but performance limitations still exist. The results demonstrate that the agentic workflow is effective and our benchmark is challenging. 
\end{itemize}

\section{Related Work}

\subsection{Multimodal Browsing Agents}

Early systems built upon LLMs demonstrated the feasibility of tool-augmented web navigation, employing search engines and web browsing tools to retrieve and integrate textual information in response to user queries \cite{searchr1, browseragent, webdancer,team2025tongyi,chang2025grail}. These agent systems established the foundational architecture for iterative retrieval and reasoning, but remained fundamentally constrained by their inability to perceive and process visual content, reducing rich multimodal web environments to text-only representations.

The advent of MLLMs drives a paradigm shift toward multimodal browsing. MMSearch-R1 \cite{mmsearchr1} is the first end-to-end reinforcement learning framework that drives MLLMs to perform multi-turn searches on demand in real-world internet environments. WebWatcher \cite{webwatcher} can leverage more external tools and design a data synthesis pipeline to produce high-quality multimodal data for training. DeepMMSearch-R1 \cite{deepmmsearch}, Skywork-R1V4 \cite{skywork} and SenseNova-MARS \cite{sensenova} introduce the image cropping tool and enhance the reasoning ability for query images. Despite their strong performance on current multimodal deep research benchmarks, a rigorous benchmark dedicated to fine-grained image understanding is still missing.

\subsection{Multimodal Browsing Benchmarks}

Multimodal browsing benchmarks are crucial for evaluating the performance of multimodal browsing agents, but existing benchmarks still have limitations. MMSearch \cite{mmsearch} pioneers the evaluation of multimodal browsing agents by introducing visual content into the search queries but the semantic information of query images can be extracted through reverse image search tool, reducing visual reasoning to text-only tool invocation. Furthermore, the problem complexity remains constrained, with most instances resolvable within limited interaction rounds that do not test sustained multi-hop reasoning capability. BrowseComp-VL \cite{webwatcher} extends this paradigm by increasing the required search iterations, thereby amplifying the depth of information retrieval. VDR-Bench \cite{vdr} employs a multi-round cropping strategy to avoid the model taking text shortcuts and using the entire image for searching. MMSearch-Plus \cite{mmsearch-plus} and MM-BrowseComp \cite{mmbc} require the agents to think with the query image and to search using fine-grained visual features, but the core evidence in the search process often includes videos or special files that cannot be accessed from publicly available sources by using tools.

To address these limitations, we propose VisBrowse-Bench, a benchmark constructed by human experts from publicly available data. This benchmark ensures that visual evidence is incorporated into the search process and activates the model's visual reasoning capabilities.

\begin{figure*}[t]
    \centering
    \includegraphics[width=\linewidth]{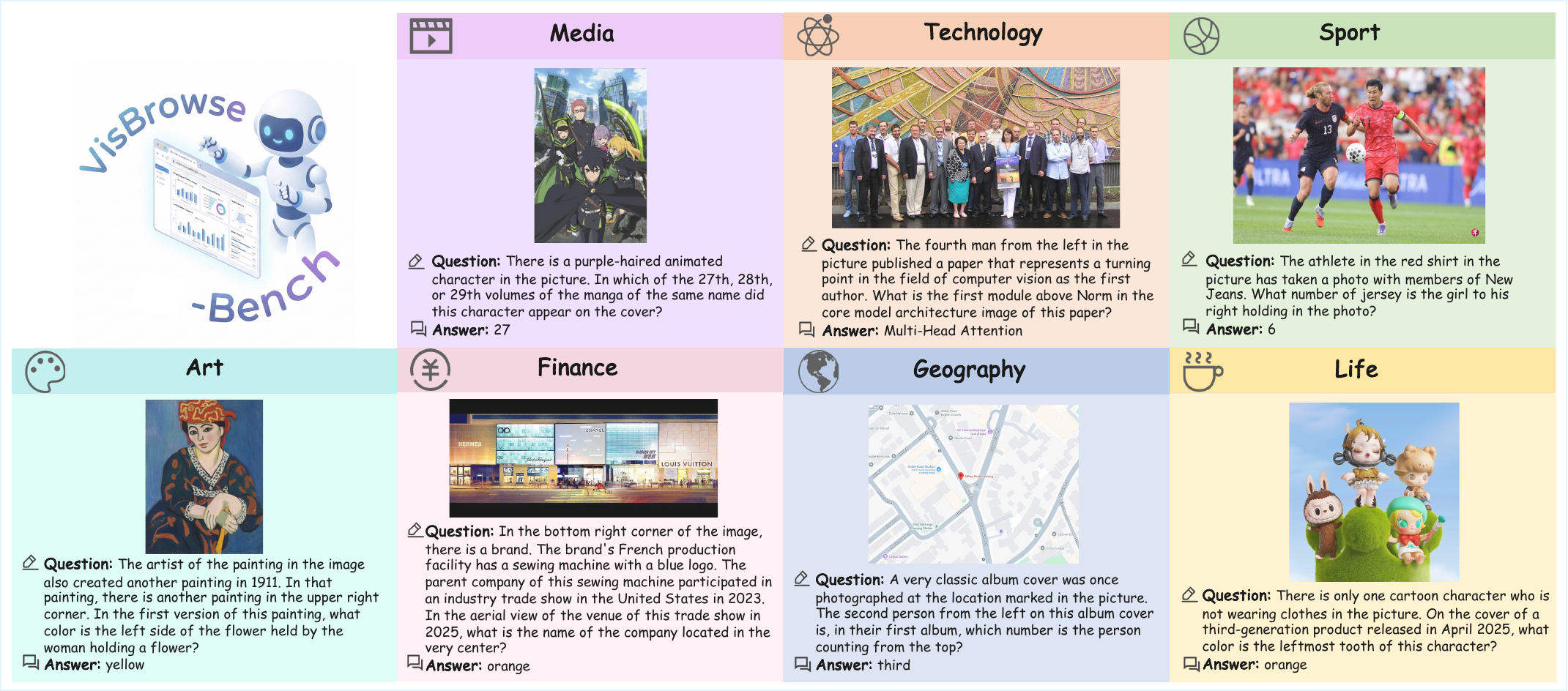}
    \caption{\textbf{Overview of the VisBrowse-Bench.} This figure shows the example questions and answers across seven categories.}
    \label{fig3}
\end{figure*}

\section{VisBrowse-Bench}

Our benchmark focuses on seven high-interest categories: Media, Life, Art, Geography, Technology, Sport and Finance, and each category has several subcategories . 
Examples of each category in VisBrowse-Bench are shown in Figure \ref{fig3}.
To ensure high-quality and data diversity, the question and answer pairs in each category were obtained through collaborative work by two experts in the field. Each question is an image–text bundle: a textual query paired with a small set of images containing initial entity information.  

In this section, we will outline the details of VisBrowse-Bench. The entire pipeline includes the development of data design criteria, data collection, and a strict validation process.

\subsection{Data Design Principles}

Our goal is to jointly evaluate the agent’s capabilities in deep search and thinking-with-image, requiring the agent to search for specific visual information during the process and reason with the visual information to obtain evidence. This emphasizes that two key principles must be included in the data design process: 

\paragraph{Multimodal Information Integration:} 
To evaluate genuine multi-modal deep search capabilities, we enforce that both the query formulation and evidence acquisition phases require active processing of visual and textual modalities. Each question is paired with a set of reference images containing task-critical cues, ensuring that query understanding necessitates visual perception. And the search space comprises interleaved text documents and visual assets. Successful evidence gathering requires the model to issue multi-modal queries and rank mixed-modality results.

\paragraph{Visual Competency Enforcement:} 
All visual information is designed to be structurally indispensable, and it cannot be substituted by textual captions or semantic descriptions, compelling the model to invoke core visual competencies, including spatial grounding, attribute perception, and relational parsing. The problem uses spatial information to describe visual elements, requiring the model to perform precise visual localization. Furthermore, the visual information involves perceptual attributes missing from the text, prompting the model to perform pixel-level analysis. The final information integration for the answer requires cross-image reasoning, which necessitates visual alignment rather than text matching.

\begin{figure*}[htbp]
    \centering
    \begin{minipage}{0.5\textwidth}
        \centering
        \subfigure[]{\includegraphics[width=\linewidth]{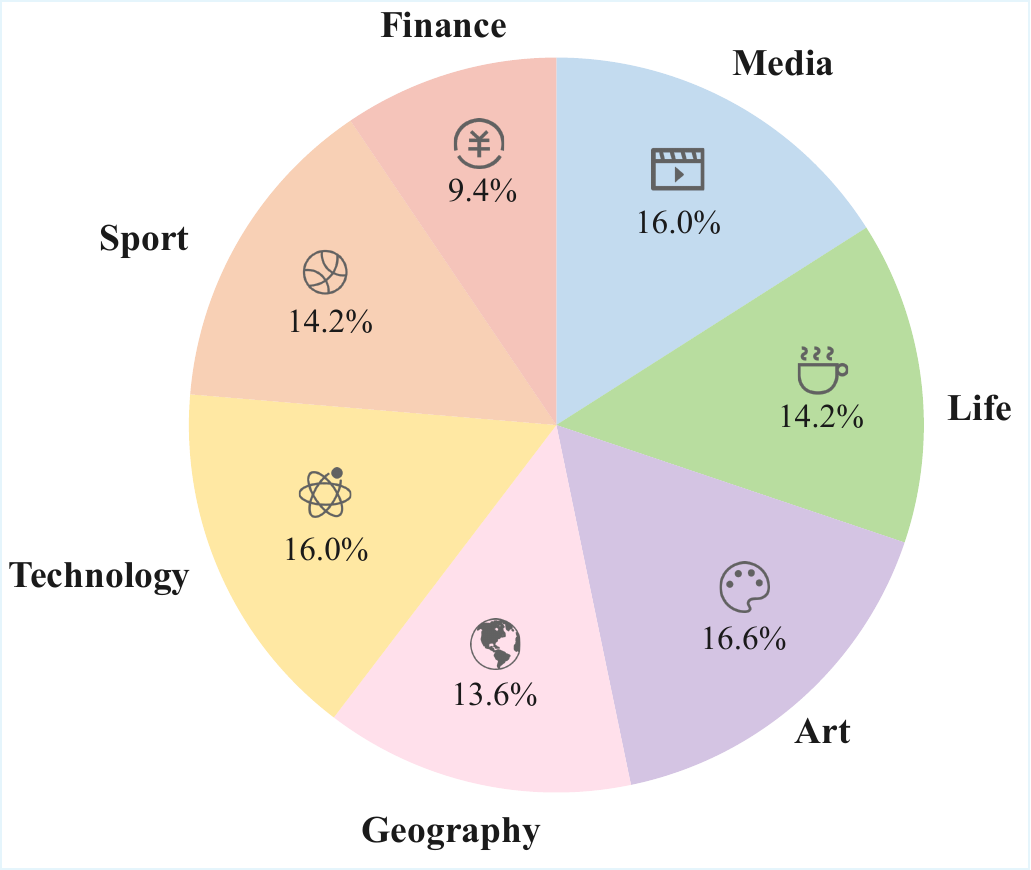}}
    \end{minipage}
    \begin{minipage}{0.48\textwidth}
        \begin{subfigure}[]{
            \begin{tabular}{lc}
                \toprule
                \textbf{Statistic} & \textbf{Number} \\
                \midrule
                Total questions & 169 \\
                Main categories & 7 \\
                Secondary categories & 24 \\
                \midrule
                Unique images & 178 \\
                Unique questions & 177 \\
                Unique answers & 177 \\
                \midrule
                Maximum question length & 94.0 \\
                Average question length & 47.7 \\
                Maximum answer length & 8 \\
                Average answer length & 1.5 \\
                \bottomrule
            \end{tabular}
            }
        \end{subfigure}
    \end{minipage}
    \caption{(a) Distribution of categories in VisBrowse-Bench. (b) Summary of statistics.}
    \label{fig4}
\end{figure*}

\subsection{Data Collection}

Two experts in each predefined category are independently responsible for selecting initial entities and managing related search keywords. These entities undergo rigorous selection to ensure a sufficient number of publicly available multimodal documents. For each initial entity, the experts initiate a systematic search process from publicly accessible data sources to obtain images that meet the above principles, especially those containing visually ambiguous entities, overlapping visual regions, or multiple co-existing entities, which require precise spatial localization and pixel-level analysis for proper interpretation.

The collection workflow proceeds through an iterative entity-event-visual information traversal. The experts first retrieve real-world events associated with the identified entity, then locate complementary visual information through these events. And the visual information must implicate a distinct related entity, distinct from the previous hop. This process of visual retrieval, event bridging, and cross-modal entity transition repeats recursively to construct multi-hop evidence chains with no fewer than two visual evidences, ensuring that both the query formulation and evidence acquisition phases require active processing of visual and textual modalities. The accumulated evidence chains undergo expert synthesis into unified questions through careful annotation, with each question designed to require integration of all retrieved evidence and explicit grounding of intermediate reasoning steps in specific visual and textual blocks, thereby guaranteeing complete instantiation of multimodal integration and visual competency enforcement principles across the entire deep search trajectory.

\subsection{Quality Validation}

Our quality validation ensures strict adherence to both design principles through a multi-layer verification framework. At the instance level, each collected question undergoes the review of two experts to confirm that visual information is structurally indispensable. We additionally enforce that the solution of each question must span a minimum of three hops and at least two visual evidence blocks, with traversal checks ensuring no single-hop shortcut can yield correct answers. 

For corpus-level quality assurance, we implement solvability and answer uniqueness verification procedures. Each question is independently solved by two domain experts with full reasoning traces documented to ensure that sufficient information exists for deterministic resolution. Instances where experts fail to reach consistent answers or identify critical evidence gaps are flagged for reconstruction or elimination.

\subsection{Dataset Statistics}

Figure \ref{fig4} details the structural properties of the VisBrowse-Bench, including the task distribution across each main category and the distribution of question length. The dataset achieves a balanced distribution across the seven main categories.
In addition, the dataset is divided into 24 secondary categories, and each question instance contains at least one image as a visual query. We compute the word-length distribution of the text portion in the question instance. The average length of the questions is 47.7 words.

\subsection{Agentic Framework}
\label{sec:framework}

To instantiate the multimodal deep search paradigm, we design a comprehensive agentic workflow equipped with five core tools: text search, image search, reverse image search, image crop and webpage visit. This agentic workflow enables the MLLMs to interact with real-world web environments through iterative tool invocation, dynamically acquiring and integrating heterogeneous evidence across modalities. The detail of agentic workflow is shown in Figure \ref{fig2}. The workflow operates as a closed loop: the MLLM parses the multimodal queries to formulate initial search strategies, issues appropriate tool calls based on current evidence gaps, processes returned results to update its internal state, and refines subsequent queries until sufficient evidence is gathered for answer synthesis. This design ensures that visual and textual evidence acquisition are interleaved throughout the reasoning trajectory, preventing premature collapse into unimodal search patterns. The tool's detailed description is as follows:

\begin{itemize}
    \item \texttt{text\_search}: The agent sends a natural language query, the SerperAPI \footnote{https://serper.dev} returns information about the ranked candidate web pages (including page titles, snippets, and URLs). 
    \begin{equation}
        text\_search(query) \rightarrow [(title, webpage\_url, snippet), ...]
    \end{equation}
    
    \item \texttt{image\_search}: The agent can use this tool to retrieve missing visual evidence during the reasoning process. Given a natural language query, SerperAPI returns semantically similar images along with the webpage information containing those images.
    \begin{equation}
        image\_search(query) \rightarrow [(title, image\_url, webpage\_url), ...]
    \end{equation}

    \item \texttt{reverse\_image\_search}: Given an image query, SerperAPI returns similar images along with the webpage information containing those images. 
    \begin{equation}
        reverse\_image\_search(image\_url) \rightarrow [(title, image\_url, webpage\_url), ...]
    \end{equation}
    
    \item \texttt{image\_crop}: Extracting refined regions of interest to support local visual reasoning and querying.
    \begin{equation}
        crop\_image(image\_url) \rightarrow cropped\_url
    \end{equation}

    \item \texttt{webpage\_visit}: The agent provides a target URL and a specific query, and JinaAPI \footnote{https://jina.ai} returns structured webpage information. The workflow uses an LLM to compress and reason about the information and generate phased conclusions relevant to the query.
    \begin{equation}
        webpage\_visit(webpage\_url, query) \rightarrow conclusion
    \end{equation}

\end{itemize}

\subsection{Discussion}

As shown in Table \ref{tab: compare}, compared to representative multimodal deep search benchmarks, VisBrowse-Bench introduces several key advancements that collectively enhance the evaluation of multimodal reasoning and visual-aware search capabilities. VisBrowse-Bench is an important benchmark for evaluating visual-native search. Specifically, VisBrowse-Bench incorporates visual evidence into the reasoning and searching process and requires MLLMs to think with images and reason across multiple images. Furthermore, our data has been verified by experts, and we guarantee that the evidence required for the searching process is publicly accessible.

\newcolumntype{L}{>{\raggedright\arraybackslash}m{5.8cm}}
\newcolumntype{C}{>{\centering\arraybackslash}m{2.2cm}}
\newcommand{\greencheck}{\textcolor{green!70!black}{\ding{51}}}
\newcommand{\redxmark}{\textcolor{red}{\ding{55}}}

\begin{table*}[t]
\centering
\normalsize
\caption{Comparison between our benchmark and other multimodal deep search benchmarks.}
\resizebox{\textwidth}{!}{
\begin{tabular}{LCCCCCCC}
\toprule
\textbf{Benchmarks} & \textbf{Multimodal queries} & \textbf{Multi-turn interaction} & \textbf{Public search information} & \textbf{Expert validation} & \textbf{Thinking with image} & \textbf{Visual evidence}  & \textbf{Cross-image reasoning}  \\
\midrule
MMSearch \cite{mmsearch} & \greencheck & \redxmark & \greencheck & \greencheck & \redxmark & \redxmark  & \redxmark  \\
BrowseComp-VL \cite{webwatcher} & \greencheck & \greencheck & \greencheck & \redxmark & \redxmark & \redxmark & \redxmark  \\
MM-BrowseComp \cite{mmbc} & \greencheck & \greencheck & \redxmark & \redxmark & \greencheck & \greencheck & \redxmark \\
MMSearch-Plus \cite{mmsearch-plus} & \greencheck & \greencheck & \redxmark & \redxmark & \greencheck & \redxmark & \redxmark \\
VDR-Bench \cite{vdr} & \greencheck & \greencheck & \greencheck & \greencheck & \greencheck & \redxmark & \redxmark \\
\midrule
\textbf{VisBrowse-Bench (Ours)} & \greencheck & \greencheck & \greencheck & \greencheck & \greencheck & \greencheck & \greencheck  \\
\bottomrule
\end{tabular}
}
\label{tab: compare}
\end{table*}

\section{Experiments}

\subsection{Experimental Setups}

\paragraph{Evaluated Models.} We evaluate both closed-source MLLMs, open-source MLLMs and Deep Research Models on VisBrowse-Bench. Closed-source models include Gemini family: Gemini-3.0-Pro, Gemini-3.0-Flash, Gemini-2.5-Pro and Gemini-2.5-Flash \cite{gemini}; GPT family: GPT-5.2 and GPT-5.1 \cite{gpt5};  Claude family: Claude-4.6-Opus and Claude-4.6-Sonnet \cite{claude}; Kimi-K2.5 \cite{kimi} and Qwen3-VL-Plus \cite{qwen3}. Open-source model includes Qwen3-VL-235B-A22B-Instruct \cite{qwen3}. Deep Research model includes o3-Deep-Research \cite{o3}.

\paragraph{Implement Details.} To quantify the impact of tool usage on performance, we evaluate each model under three progressively enhanced tool-use methods:
\begin{itemize}
    \item \textbf{Direct Answer:} Models answer the question relying on internal parametric knowledge, without external tool access.
    \item \textbf{+ Text Search (+ TS):} Models can only use \texttt{text\_search} and \texttt{webpage\_visit} tools to support evidence acquisition.
    \item \textbf{+ Image Search (+ IS):} Models can use all the tools in the Section \ref{sec:framework} framework to collect visual and textual evidence.
\end{itemize}

\paragraph{Evaluation Metrics.} We use accuracy (\%) as the metric to evaluate the model's performance on VisBrowse-Bench. 
Through the regular expression matching method, we extract the model's final answer. 
We use GPT-5.1 as the judge model and employ the LLM-as-Judge method to compare the model's answer with the ground truth and determine its correctness. Details of the judgment model's prompt are in the Appendix \ref{sec: prompt}.

\begin{table*}[!ht]
\scriptsize
\centering
\caption{Main results of the VisBrowse-Bench across different categories. The evaluation metric is accuracy (\%). Each model is evaluated using three methods: Direct answer, + TS and + IS. The \textcolor{green!70!black}{green numbers} represent performance improvement compared to the previous method, and \textcolor{red}{red numbers} represent performance degradation. The \textbf{bold numbers} represent the best accuracy overall or in each category, and \underline{underlined numbers} represent the second-best accuracy.}
\label{tab:main}
\begin{tabular}{lllccccccc}
\toprule
\multicolumn{1}{c}{\multirow{2}{*}{\textbf{Model}}} & \multicolumn{1}{c}{\multirow{2}{*}{\textbf{Method}}} & \multicolumn{1}{c}{\multirow{2}{*}{\textbf{Overall}}} &       &      &     & \textbf{Category}  &           &       &         \\ \cmidrule(lr){4-10}
\multicolumn{1}{c}{}                       & \multicolumn{1}{c}{}                        & \multicolumn{1}{c}{}                      & Meida & Life & Art & Geography & Techology & Sport & Finance \\
\midrule
\rowcolor{gray!20} \multicolumn{10}{c}{\textbf{Closed-source Model}} \\
\midrule
\multicolumn{1}{l|}{\multirow{3}{*}{\textbf{Gemini-3.0-Pro}}} & Direct Answer & 23.7 & 25.9 & 16.7 & 28.6 & 30.4 & 25.9 & 16.7 & 18.8 \\
\multicolumn{1}{l|}{}         & + TS & 38.5 \textcolor{green!70!black}{\scriptsize \textit{\textbf{(+14.8)}}} & 44.4 & \underline{45.8} & \underline{35.7} & 39.1 & 25.9 & \textbf{50.0} & \underline{25.0} \\
\multicolumn{1}{l|}{}         & + IS & 40.2 \textcolor{green!70!black}{\scriptsize \textit{\textbf{(+1.7)}}} & 44.4 & \underline{45.8} & \textbf{42.9} & 43.5 & 48.1 & 29.2 & 18.8 \\
\midrule
\multicolumn{1}{l|}{\multirow{3}{*}{\textbf{Gemini-3.0-Flash}}} & Direct Answer & 32.5 & 29.6 & 33.3 & 28.6 & 34.8 & 37.0 & 33.3 & \textbf{31.2} \\
\multicolumn{1}{l|}{}         & + TS & 37.9 \textcolor{green!70!black}{\scriptsize \textit{\textbf{(+5.4)}}} & 40.7 & \textbf{50.0} & 28.6 & 43.5 & 29.6 & \underline{45.8} & \underline{25.0} \\
\multicolumn{1}{l|}{}         & + IS & 39.1 \textcolor{green!70!black}{\scriptsize \textit{\textbf{(+1.2)}}} & 40.7 & \underline{45.8} & 32.1 & 43.5 & 37.0 & \underline{45.8} & \underline{25.0} \\
\midrule
\multicolumn{1}{l|}{\multirow{3}{*}{\textbf{Gemini-2.5-Pro}}}  & Direct Answer & 19.5 & 25.9 & 4.2 & 28.6 & 26.1 & 14.8 & 12.5 & \underline{25.0} \\
\multicolumn{1}{l|}{}         & + TS & 20.7 \textcolor{green!70!black}{\scriptsize \textit{\textbf{(+1.2)}}} & 40.7 & 4.2 & 25.0 & 26.1 & 18.5 & 16.7 & 6.2 \\
\multicolumn{1}{l|}{}         & + IS & 26.6 \textcolor{green!70!black}{\scriptsize \textit{\textbf{(+5.9)}}} & 37.0 & 25.0 & 32.1 & 34.8 & 22.2 & 25.0 & 0.0 \\
\midrule
\multicolumn{1}{l|}{\multirow{3}{*}{\textbf{Gemini-2.5-Flash}}}  & Direct Answer & 9.5 & 14.8 & 4.2 & 0.0 & 13.0 & 22.2 & 8.3 & 0.0 \\
\multicolumn{1}{l|}{}         & + TS & 17.2 \textcolor{green!70!black}{\scriptsize \textit{\textbf{(+7.7)}}} & 40.7 & 16.7 & 10.7 & 21.7 & 7.4 & 12.5 & 6.2 \\
\multicolumn{1}{l|}{}         & + IS & 20.7 \textcolor{green!70!black}{\scriptsize \textit{\textbf{(+3.5)}}} & 37.0 & 12.5 & 14.3 & 26.1 & 29.6 & 16.7 & 0.0 \\
\midrule
\multicolumn{1}{l|}{\multirow{3}{*}{\textbf{GPT-5.2}}} & Direct Answer & 14.8 & 7.4 & 25.0 & 21.4 & 13.0 & 18.5 & 12.5 & 0.0 \\
\multicolumn{1}{l|}{}         & + TS & 26.0 \textcolor{green!70!black}{\scriptsize \textit{\textbf{(+11.2)}}} & 33.3 & 41.7 & 17.9 & 26.1 & 25.9 & 25.0 & 6.2 \\
\multicolumn{1}{l|}{}         & + IS &  28.4 \textcolor{green!70!black}{\scriptsize \textit{\textbf{(+2.4)}}} & 37.0 & 29.2 & 25.0 & 30.4 & 37.0 & 25.0 & 6.2 \\
\midrule
\multicolumn{1}{l|}{\multirow{3}{*}{\textbf{GPT-5.1}}}  & Direct Answer & 13.0 & 14.8 & 20.8 & 21.4 & 17.4 & 11.1 & 0.0 & 0.0 \\
\multicolumn{1}{l|}{}         & + TS & 16.6 \textcolor{green!70!black}{\scriptsize \textit{\textbf{(+3.6)}}} & 25.9 & 29.2 & 14.3 & 21.7 & 14.8 & 0.0 & 6.2 \\
\multicolumn{1}{l|}{}         & + IS & 23.1 \textcolor{green!70!black}{\scriptsize \textit{\textbf{(+6.5)}}} & 22.2 & 41.7 & 14.3 & 30.4 & 22.2 & 12.5 & 18.8 \\
\midrule
\multicolumn{1}{l|}{\multirow{3}{*}{\textbf{Claude-4.6-Opus}}}  & Direct Answer & 27.2 & 37.0 & 25.0 & 28.6 & 26.1 & 29.6 & 20.8 & 18.8 \\
\multicolumn{1}{l|}{}         & + TS & 42.6 \textcolor{green!70!black}{\scriptsize \textit{\textbf{(+15.4)}}} & 48.1 & 45.8 & \underline{35.7} & 56.5 & 48.1 & 33.3 & \underline{25.0} \\
\multicolumn{1}{l|}{}         & + IS & \textbf{47.6} \textcolor{green!70!black}{\scriptsize \textit{\textbf{(+5.0)}}} & 53.6 & \textbf{50.0} & \underline{35.7} & \textbf{56.5} & \textbf{59.3} & \underline{45.8} & \underline{25.0} \\
\midrule
\multicolumn{1}{l|}{\multirow{3}{*}{\textbf{Claude-4.6-Sonnet}}}  & Direct Answer & 13.0 & 18.5 & 12.5 & 14.3 & 26.1 & 11.1 & 4.2 & 0.0 \\
\multicolumn{1}{l|}{}         & + TS & 23.7 \textcolor{green!70!black}{\scriptsize \textit{\textbf{(+10.7)}}} & 29.6 & 20.8 & 25.0 & 39.1 & 14.8 & 20.8 & 12.5 \\
\multicolumn{1}{l|}{}         & + IS & 18.3 \textcolor{red}{\scriptsize \textit{\textbf{(-5.4)}}} & 25.9 & 20.8 & 3.6 & 30.4 & 18.5 & 20.8 & 6.2 \\
\midrule
\multicolumn{1}{l|}{\multirow{3}{*}{\textbf{Kimi-K2.5}}}  & Direct Answer & 21.3 & 18.5 & 12.5 & 17.9 & 26.1 & 25.9 & 29.2 & 18.8 \\
\multicolumn{1}{l|}{}         & + TS & 21.9 \textcolor{green!70!black}{\scriptsize \textit{\textbf{(+0.6)}}} & 29.6 & 16.7 & 14.3 & 34.8 & 29.6 & 16.7 & 6.2 \\
\multicolumn{1}{l|}{}         & + IS & \underline{41.4} \textcolor{green!70!black}{\scriptsize \textit{\textbf{(+19.5)}}} & \textbf{63.0} & \textbf{50.0} & 25.0 & 39.1 & \underline{51.9} & 33.3 & 18.8 \\
\midrule
\multicolumn{1}{l|}{\multirow{3}{*}{\textbf{Qwen3-VL-Plus}}}  & Direct Answer & 17.8 & 18.5 & 20.8 & 17.9 & 8.7 & 25.9 & 20.8 & 6.2 \\
\multicolumn{1}{l|}{}         & + TS & 27.8 \textcolor{green!70!black}{\scriptsize \textit{\textbf{(+10.0)}}} & 33.3 & 29.2 & 21.4 & 30.4 & 37.0 & 20.8 & 18.8  \\
\multicolumn{1}{l|}{}         & + IS & 32.5 \textcolor{green!70!black}{\scriptsize \textit{\textbf{(+4.7)}}} & 37.0 & \textbf{50.0} & 25.0 & 39.1 & 40.7 & 16.7 & 12.5 \\
\midrule
\rowcolor{gray!20} \multicolumn{10}{c}{\textbf{Open-source Model}} \\
\midrule
\multicolumn{1}{l|}{\multirow{3}{*}{\textbf{Qwen3-VL-235B-A22B}}}  & Direct Answer & 10.1 & 11.1 & 20.8 & 3.6 & 8.7 & 14.8 & 8.3 & 0.0 \\
\multicolumn{1}{l|}{}         & + TS & 11.2 \textcolor{green!70!black}{\scriptsize \textit{\textbf{(+1.0)}}} & 11.1 & 12.5 & 0.0 & 17.4 & 22.2 & 12.5 & 0.0 \\
\multicolumn{1}{l|}{}         & + IS & 14.2 \textcolor{green!70!black}{\scriptsize \textit{\textbf{(+3.0)}}} & 25.9 & 25.0 & 3.6 & 26.1 & 11.1 & 4.2 & 0.0 \\
\midrule
\rowcolor{gray!20} \multicolumn{10}{c}{\textbf{Deep Research Model}} \\
\midrule
\multicolumn{1}{l|}{\textbf{o3-Deep-Research}}  & \multirow{1}{*}{Direct Answer} & 41.1 & \underline{55.6} & 41.7 & 21.4 & \underline{52.2} & 48.1 & 37.5 & \underline{25.0} \\
\bottomrule
\end{tabular}
\end{table*}

\subsection{Results and Analysis}

The main results on VisBrowse-Bench are shown in Table \ref{tab:main}, and we make the analysis as follow:

\paragraph{Inherent challenge.} Under the direct answer method, all models perform poorly, demonstrating the challenging nature of our benchmark in complex real-world search tasks, where the model's parameterized knowledge is insufficient to handle complex queries requiring dynamically acquired evidence. 

\paragraph{Insufficiency of textual evidence.} Introducing text search (+ TS) improves the performance of all models to varying degrees, indicating that the models can correctly interact with the tool and effectively acquire information. However, the improvement is limited for most models, demonstrating that the visual evidence needed to answer questions in our benchmark cannot be easily obtained through text retrieval alone. 

\paragraph{Necessity of visual evidence.} The integration of image search (+ IS) further improves the performance of models, demonstrating the effectiveness of our framework in driving the model to actively discover the visual evidence needed to solve the problem. However, the Claude-4.6-Sonnet cannot use image search tools correctly, which causes the framework to fail and results in lower accuracy.


On our benchmark, the best-performing model achieves only 47.6\% accuracy, while the top open-source model reaches merely 14.2\%, reflecting the inherent difficulty of our dataset. Additionally, we find that these models with rich prior knowledge and strong reasoning capabilities tend to rely solely on text-based reasoning. Case studies are included in the Appendix \ref{sec: case}.

\subsection{Analysis of Tool Usage}

\begin{figure*}[t]
    \centering
    \includegraphics[width=
    \linewidth]{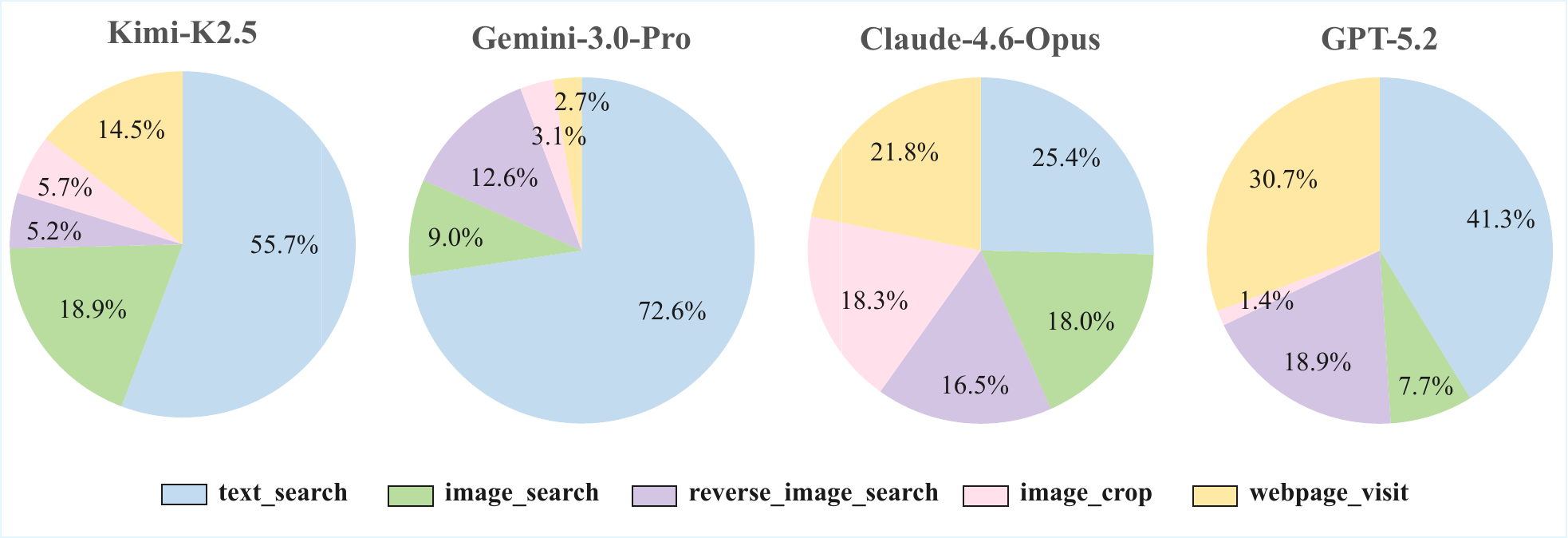}
    \caption{The percentage of times each of the five tools was used in four MLLMs: Kimi-K2.5, Gemini-3.0-Pro, Claude-4.6-Opus and GPT-5.2}
    \label{fig5}
\end{figure*}

To analyze the behavior and strategies of the agents in multimodal deep search, we statistically analyze the tool usage distribution of four models under + IS methods. Figure \ref{fig5} shows the usage percentage of each tool.

Through our data analysis, we observe significant variations in tool-usage preferences across different models. Claude-4.6-Opus can use the five tools in a relatively balanced way and achieves the best performance. Compared to Gemini-3.0-Pro and GPT-5.2, Kimi-K2.5 utilizes \texttt{image\_search} tool more frequently, enabling it to uncover more visual information within web pages through text queries, thus resulting in the greatest performance improvement under the + IS method.  The tool usage details for all models are shown in the Appendix \ref{sec: tool}.

\section{Conclusion}

We identify two key gaps in existing multimodal browsing agent benchmarks: they inadequately test visual reasoning and overlook multimodal information integration in the reasoning process. To address this, we introduce VisBrowse-Bench, a benchmark constructed by human experts with 169 questions covering seven categories. It is designed so that multimodal information fusion is required during search and visual understanding is essential for task completion. We implement an agentic workflow with tools including text search, image search, reverse image search, image cropping, and webpage browsing to encourage proactive collection and reasoning over visual content.

\bibliographystyle{unsrtnat} 
\bibliography{main}

\newpage
\section*{Appendix}
\label{sec:appendix}

\subsection{Prompts of agentic framework}
\label{sec: prompt}

\begin{tcolorbox}[title=System Prompt]
You are a Web Information Seeking Master. Your task is to thoroughly seek the internet for information and provide accurate answers to visual questions.

\vspace{3pt}

As you proceed, adhere to the following principles:

\vspace{3pt}

1. Decompose the original visual question into sub-questions and solve them step by step. Summarize the knowledge obtained from the previous round of dialogue, then think about what is next sub-question.

\vspace{3pt}

2. Whether you can answer the question or not, you should describe the image in detail. if the image includes multiple sub-image, you should describe each one separately.

\vspace{3pt}

3. Before calling any tools, you must provide a brief explanation of why you are calling the tool and what you expect to achieve.

\vspace{3pt}

4. You should provide the final answer within 15 turns, regardless of whether all valid information has been collected.

\end{tcolorbox}

\begin{tcolorbox}[title=Input Prompt]
You are an intelligent agent engaged in a conversation with a user. The user poses a question and provides a corresponding image for context. As an agent, you approach the problem with care and methodical precision, following a multi-step process to arrive at a solution. You utilize a variety of tools, ensuring that the information gathered from each one is cross-validated before you reach a final answer. Rather than relying on any single tool for accuracy, you employ multiple tools iteratively to prioritize the comprehensiveness and reliability of your responses.

\vspace{3pt}

To be successful, it is very important to follow the following rules:

\vspace{3pt}

1. The assistant starts with one or more cycles of (thinking about which tool to use -> performing tool call -> waiting for tool response), and ends with (thinking about the answer -> answer of the question).

2. If additional visual information is needed during the searching process, 'image\_search' tool can be used to search for images.

\vspace{3pt}

3. You can use 'crop\_image' tool to zoom in on a specific region of the image and search for it.

\vspace{3pt}

4. You can only make one tool call per round and wait for the tool's response.

\vspace{3pt}

5. Your answer should be inside '<answer></answer>' tags, and the answer must be the most concise output.

\vspace{3pt}

Input Question: {}

Input image\_url: {}

\end{tcolorbox}

\begin{tcolorbox}[title=LLM-as-Judge Prompt]
You are a rigorous question-and-answer quality assessment expert. Given a question, your task is to compare the "groud truth" and the "model answer", determine whether they are semantically equivalent, and output "yes" or "no".

\vspace{3pt}

question: {}

ground truth: {}

model answer: {}

\end{tcolorbox}

\subsection{Details of tool usage}
\label{sec: tool}

We analyze the distribution of tool invocations across the five available tools: text search (TS), image search (IS), reverse image search (RIS), image crop (IC), and webpage visit (WV). Table \ref{tab: tool} presents the tool usage frequencies for MLLMs, revealing distinct strategic signatures that correlate with performance outcomes.

The complexity of our benchmark is substantiated by the substantial tool interaction depth required for successful problem resolution, and the extended interaction sequences in our evaluation reflect the inherent multi-hop, multi-modal nature of the constructed problems. MLLMs must iteratively refine visual hypotheses through targeted image search, verify provenance via reverse image search, extract fine-grained details through cropping, and corroborate findings with textual evidence—no single tool or modality provides sufficient information for independent resolution.

\begin{table*}[!ht]

\centering
\caption{Statistic of tool usage for MLLMs.}
\label{tab: tool}
\begin{tabular}{lcccccc}
\toprule
\multicolumn{1}{c}{\multirow{2}{*}{\textbf{Model}}}  & \multicolumn{1}{c}{\multirow{2}{*}{\textbf{Average turns}}} & & & \textbf{Tool usage (\%)} & & \\ \cmidrule(lr){3-7}
 \multicolumn{1}{c}{} & \multicolumn{1}{c}{} & TS & IS & RIS & IC & WV \\
\midrule
\rowcolor{gray!20} \multicolumn{7}{c}{\textbf{Closed-source Model}} \\
\midrule
\multicolumn{1}{l|}{\textbf{Gemini-3.0-Pro}} & 12.7 & 72.6 & 9.0 & 12.6 & 3.1 & 2.7 \\

\multicolumn{1}{l|}{\textbf{Gemini-3.0-Flash}} & 9.1 & 75.3 & 15.4 & 5.2 & 0.8 & 3.3  \\

\multicolumn{1}{l|}{\textbf{Gemini-2.5-Pro}} & 6.0 & 44.4 & 20.4 & 15.9 & 5.6 & 13.7 \\

\multicolumn{1}{l|}{\textbf{Gemini-2.5-Flash}} & 5.4 & 45.4 & 11.6 & 18.1 & 3.9 & 21.0 \\

\multicolumn{1}{l|}{\textbf{GPT-5.2}} & 7.3 & 41.3 & 7.7 & 18.9 & 1.4 & 30.7 \\

\multicolumn{1}{l|}{\textbf{GPT-5.1}} & 3.5 & 45.6 & 5.9 & 31.6 & 0.7 & 16.2 \\

\multicolumn{1}{l|}{\textbf{Claude-4.6-Opus}} & 12.3 & 25.4 & 18.0 & 16.5 & 18.3 & 21.8 \\

\multicolumn{1}{l|}{\textbf{Claude-4.6-Sonnet}} & 10.9 & 28.6 & 8.3 & 29.2 & 5.0 & 28.9 \\

\multicolumn{1}{l|}{\textbf{Kimi-K2.5}} & 8.6 & 55.7 & 18.9 & 5.2 & 5.7 & 14.5 \\

\multicolumn{1}{l|}{\textbf{Qwen3-VL-Plus}} & 5.5 & 56.6 & 23.8 & 2.4 & 7.6 & 9.6 \\

\midrule
\rowcolor{gray!20} \multicolumn{7}{c}{\textbf{Open-source Model}} \\
\midrule

\multicolumn{1}{l|}{\textbf{Qwen3-VL-235B-A22B}} & 5.5 & 46.6 & 42.7 & 0.0 & 7.6 & 3.1 \\

\bottomrule
\end{tabular}
\end{table*}

\subsection{Case Study}
\label{sec: case}

\definecolor{myblue}{RGB}{68,114,194}

\definecolor{mybrown}{RGB}{127,96,0}

\subsubsection{Analysis of successful case}

we present a detailed analysis of a successful search trajectory of Claude-4.6-Opus in the Media category. Despite Claude-4.6-Opus has powerful parametric knowledge, it effectively integrates five tools to obtain evidence for cross-validation during the search and reasoning process, ultimately leading to a successful answer.

\paragraph{Question:} The person at the top right of this image starred in a critically acclaimed film in 2023. In the main visual poster of this film, the character in the center is shown having a conversation with another character by the lake in Princeton. Who played the character dressed in black in that scene?

\begin{figure}[h]
    \centering
    \includegraphics[width=0.3\linewidth]{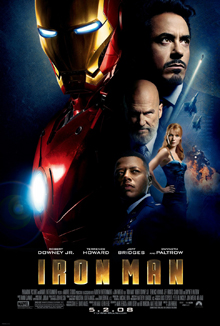}
    \caption{Visual query}
    \label{fig:placeholder}
\end{figure}

\begin{tcolorbox}[title=Trajectory of Claude-4.6-Opus \hspace{3pt} \textcolor{green!70!black}{Successful}, breakable]

\colorbox{gray!60}{\parbox{\linewidth}{\centering \textbf{Step 1}}}

\vspace{1em}

\textbf{Assistant:}

Let me analyze this image. It's the movie poster for Iron Man (2008). I need to identify the person at the top right of the image. Let me crop that area to get a better look.

\textcolor{myblue}{\textit{<tool\_call>}} \\
\textcolor{mybrown}{image\_crop([0.6,0,1,0.45])} \\
\textcolor{myblue}{\textit{</tool\_call>}}
    
\textbf{Tool:}

The url of cropped image: https://...

\vspace{1em}

\colorbox{gray!60}{\parbox{\linewidth}{\centering \textbf{Step 2}}}

\vspace{1em}

\textbf{Assistant:}

I can see Robert Downey Jr. at the top right of the image. He's the one in the suit. Now I need to find out what critically acclaimed film he starred in in 2023. That would be "Oppenheimer" (2023), directed by Christopher Nolan, where he played Lewis Strauss.

Now I need to find the main visual poster of Oppenheimer and identify the character in the center, then find the scene where that character has a conversation with another character by the lake in Princeton.

In Oppenheimer, the central character is J. Robert Oppenheimer, played by Cillian Murphy. There's a famous scene by the lake in Princeton where Oppenheimer talks with Albert Einstein. Einstein was played by Tom Conti in the film.

But let me verify this. The question asks about the "main visual poster" - the character in the center of the poster. Let me search for the Oppenheimer poster and the lake scene.

\textcolor{myblue}{\textit{<tool\_call>}} \\
\textcolor{mybrown}{text\_search("Robert Downey Jr critically acclaimed film 2023")} \\
\textcolor{myblue}{\textit{</tool\_call>}}

\textbf{Tool:}

A Google search for 'Robert Downey Jr critically acclaimed film 2023' found 5 results:

Web Results

1. [Robert Downey Jr: High Altitude (2023) - IMDb](https://www.imdb.com/...)

Charting Robert Downey Jr.'s journey from stardom to personal struggles and remarkable comeback as one of Hollywood's most charismatic actors through his iconic ...

2. [Robert Downey Jr. Says 'Oppenheimer' Is His Best Movie - YouTube](https://www.youtube.com/...)

Date published: Jul 15, 2023

Graeme O'Neil takes a listen to the crowd's reaction after Robert Downey Jr. declaring \"Oppenheimer\" is the best movie his ever been in, ...

3. [Robert Downey Jr: High Altitude (2023) Full Movie - YouTube](https://www.youtube.com/...)

Date published: Dec 14, 2023

From stardom, hitting rock bottom, and back to the top, Robert Downey Jr. is one of the most successful actors in Hollywood, thanks to his ...

4. [Robert Downey Jr. Says 'Oppenheimer' Is His Best Movie, but It's ...](https://decider.com/...)

Date published: Jul 18, 2023

Over 50 years into his career, the actor claims that his latest movie, Oppenheimer, is the “best movie” he's ever been in.

5. [Critics are calling Oppenheimer one of Robert Downey Jr.'s 'career ...](https://www.instagram.com/...)

Date published: Nov 23, 2023

Critics are calling Oppenheimer one of Robert Downey Jr.'s 'career-best performances' (The Wrap). For the first time, go behind the scenes and ...
\vspace{1em}

\colorbox{gray!60}{\parbox{\linewidth}{\centering \textbf{Step 3}}}

\vspace{1em}

\textbf{Assistant:}

Confirmed - Robert Downey Jr. starred in "Oppenheimer" (2023). Now I need to find the main visual poster of Oppenheimer, identify the central character, and then find the lake scene in Princeton.

The central character on the Oppenheimer poster is J. Robert Oppenheimer, played by Cillian Murphy. The famous lake scene in Princeton is between Oppenheimer and Albert Einstein. Let me search for this specific scene and who played Einstein.

\textcolor{myblue}{\textit{<tool\_call>}} \\
\textcolor{mybrown}{image\_search("Oppenheimer 2023 movie main poster")} \\
\textcolor{myblue}{\textit{</tool\_call>}}

\textbf{Tool:}

1. Image: https://m.media-amazon.com/..., Title: Oppenheimer (2023) - IMDb, Webpage Url: https://www.imdb.com/...

2. Image: https://m.media-amazon.com/..., Title: Amazon.com: XIHOO Oppenheimer - 2023 Movie Poster 11x17 ..., Webpage Url: https://www.amazon.com/...

3. Image: https://i.ebayimg.com/..., Title: Oppenheimer 2023 Premium Movie Poster MADE IN USA - CIN559, Webpage Url: https://www.ebay.com/...

\vspace{1em}

\colorbox{gray!60}{\parbox{\linewidth}{\centering \textbf{Step 4}}}

\vspace{1em}

\textbf{Assistant:}

The main poster of Oppenheimer shows Cillian Murphy as J. Robert Oppenheimer in the center. Now I need to find the lake scene in Princeton where Oppenheimer talks to Einstein. Let me search for this specific scene.

\textcolor{myblue}{\textit{<tool\_call>}} \\
\textcolor{mybrown}{text\_search("Oppenheimer movie lake scene Princeton Einstein who played")} \\
\textcolor{myblue}{\textit{</tool\_call>}}

\textbf{Tool:}

A Google search for 'Oppenheimer movie lake scene Princeton Einstein who played' found 5 results:

Web Results

1. [Two great minds. Christopher Nolan shot scenes for Oppenheimer ...](https://www.facebook.com/...)

Date published: Jul 11, 2023

Christopher Nolan shot scenes for Oppenheimer at Princeton University, where Oppenheimer and Einstein worked together after World War II. No ...

2. [Was The Oppenheimer and Einstein Scene In Christopher Nolans ...](https://www.reddit.com/...)

Date published: Jan 4, 2024

One specific scene, where Oppie asks Einstein about the math that an atomic explosion may destroy the world, may have happened but was not documented.

3. [Albert Einstein in Oppenheimer and Prisoner in The Dark Knight Rises](https://www.tiktok.com/...)

Date published: Jul 24, 2023

He also plays the prisoner in The Dark Knight Rises. who tells Bruce Wayne, rise. I love the fact that in both films. he has a significant ...

4. ['Oppenheimer' behind the scenes in N.J.](https://www.nj.com/...)

Date published: Mar 11, 2024

Location manager breaks down shooting in Einstein's real office and how Princeton can be like Times Square.

5. [Who Plays Albert Einstein in Christopher Nolan's 'Oppenheimer'?](https://collider.com/...)

Date published: May 19, 2024

In Oppenheimer, Albert Einstein is played by Tom Conti, who comes to the latest Nolan event following a long and lauded career.

\vspace{1em}

\colorbox{gray!60}{\parbox{\linewidth}{\centering \textbf{Step 5}}}

\vspace{1em}

\textbf{Assistant:}

I found that Albert Einstein in Oppenheimer is played by Tom Conti. Now let me verify the lake scene. The question asks about the character "dressed in black" in the lake scene in Princeton. Let me search for images of this specific scene to confirm who is dressed in black.

\textcolor{myblue}{\textit{<tool\_call>}} \\
\textcolor{mybrown}{image\_search("Oppenheimer Einstein lake scene Princeton conversation")} \\
\textcolor{myblue}{\textit{</tool\_call>}}

\textbf{Tool:}

1. Image: https://hips.hearstapps.com/..., Title: So Did Oppenheimer Really Meet Einstein?, Webpage Url: https://www.esquire.com/...

2. Image: https://lookaside.instagram.com/..., Title: In the comments, caption this photo of Oppenheimer and ..., Webpage Url: https://www.instagram.com/...

3. Image: https://lookaside.instagram.com/..., Title: Seen the movie Oppenheimer? It contains several scenes ..., Webpage Url: https://www.instagram.com/...

\vspace{1em}

\colorbox{gray!60}{\parbox{\linewidth}{\centering \textbf{Step 6}}}

\vspace{1em}

\textbf{Assistant:}

Let me look at the lake scene image more closely to see who is dressed in black.

\textcolor{myblue}{\textit{<tool\_call>}} \\
\textcolor{mybrown}{webpage\_visit(goal="Find details about the lake scene between Oppenheimer and Einstein in Princeton, and who is dressed in black", url="https://www.esquire.com/...")} \\
\textcolor{myblue}{\textit{</tool\_call>}}

\textbf{Tool:}

The useful information in https://www.esquire.com/... for user goal Find details about the lake scene between Oppenheimer and Einstein in Princeton, and who is dressed in black as follows: 

Evidence in page: 

...

Summary: 

...

\vspace{1em}

\colorbox{gray!60}{\parbox{\linewidth}{\centering \textbf{Step 7}}}

\vspace{1em}

\textbf{Assistant:}

Let me search for images of the specific Oppenheimer and Einstein scene by the lake to see who is dressed in black. The famous scene takes place by a pond/lake at the Institute for Advanced Study in Princeton.

\textcolor{myblue}{\textit{<tool\_call>}} \\
\textcolor{mybrown}{image\_search("Oppenheimer Einstein pond scene Tom Conti Cillian Murphy")} \\
\textcolor{myblue}{\textit{</tool\_call>}}

\textbf{Tool:}

1. Image: https://hips.hearstapps.com/..., Title: So Did Oppenheimer Really Meet Einstein?, Webpage Url: https://www.esquire.com/...

2. Image: https://i2-prod.dailyrecord.co.uk/..., Title: Albert Einstein in Oppenheimer brought to life by veteran ..., Webpage Url: https://www.scottishdailyexpress.co.uk/...

3. Image: https://blogger.googleusercontent.com/..., Title: Highly Reasonable: Doesn't That Pond Look Great?, Webpage Url: http://highlyreasonable.blogspot.com/..

...

...

...

\vspace{1em}

\colorbox{gray!60}{\parbox{\linewidth}{\centering \textbf{Step 15}}}

\vspace{1em}

\textbf{Assistant:}

Looking at the image from the lake/pond scene in Oppenheimer, I can see:

- On the left: Tom Conti as Albert Einstein, wearing a dark grey/charcoal overcoat (which could be described as dark/black) with white hair and holding a black hat

- On the right: Cillian Murphy as J. Robert Oppenheimer, wearing a brown coat and brown hat

The question asks about "the character dressed in black in that scene." Einstein (Tom Conti) is wearing the darker clothing - a dark charcoal/grey-black overcoat and holding a black hat. Oppenheimer is in brown.

So the character dressed in black (or dark clothing) in the lake scene is Albert Einstein, played by Tom Conti.

Based on my research:

1. **Person at top right of the Iron Man poster**: Robert Downey Jr.

2. **Critically acclaimed 2023 film**: *Oppenheimer* (directed by Christopher Nolan)

3. **Character in the center of the Oppenheimer poster**: J. Robert Oppenheimer (played by Cillian Murphy)

4. **Lake scene in Princeton**: Oppenheimer has a pivotal conversation with Albert Einstein by the pond at the Institute for Advanced Study in Princeton.

5. **Character dressed in black/dark clothing**: In the scene, Einstein is wearing a dark charcoal/black overcoat and holding a black hat, while Oppenheimer wears a brown coat.

Albert Einstein was played by **Tom Conti**.

\textcolor{green!70!black}{<answer>Tom Conti</answer>}

\end{tcolorbox}

\subsubsection{Analysis of failed case}

In this example, although the model successfully finds key visual evidence through interaction with the tool, it dose not use its powerful visual reasoning ability to extract useful information from the evidence. Instead, it incorrectly invoke the visit tool to attempt to analyze the image.

\paragraph{Question:} The brand of the gray SUV in the upper left corner of the image launched a new collection in August 2023. What color tie is the brand ambassador wearing in the poster for his 2025 world tour?

\begin{figure}[h]
    \centering
    \includegraphics[width=0.4\linewidth]{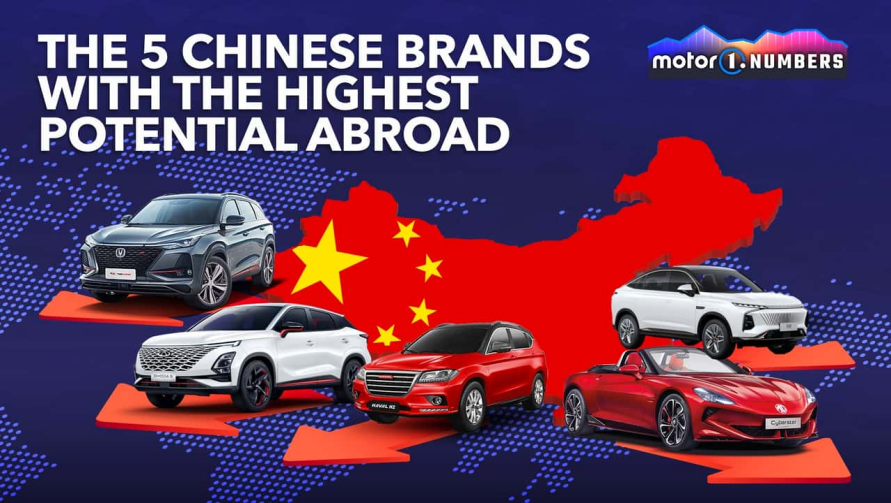}
    \caption{Visual query}
    \label{fig:placeholder}
\end{figure}

\begin{tcolorbox}[title=Trajectory of Claude-4.6-Opus \hspace{10pt} \textcolor{red}{Failed}, breakable]

\colorbox{gray!60}{\parbox{\linewidth}{\centering \textbf{Step 1}}}

\vspace{1em}

\textbf{Assistant:}

Let me analyze the image to identify the gray SUV in the upper left corner. I need to zoom in on that area to get a better look at the brand.

\textcolor{myblue}{\textit{<tool\_call>}} \\
\textcolor{mybrown}{image\_crop([0,0.15,0.35,0.75])} \\
\textcolor{myblue}{\textit{</tool\_call>}}

\textbf{Tool:}

The url of cropped image: https://...

\vspace{1em}

\colorbox{gray!60}{\parbox{\linewidth}{\centering \textbf{Step 2}}}

\vspace{1em}

\textbf{Assistant:}

I can see the gray SUV in the upper left corner has a logo that looks like the Changan logo - a "V" shape or inverted triangle. Let me zoom in more on the logo to confirm.

\textcolor{myblue}{\textit{<tool\_call>}} \\
\textcolor{mybrown}{image\_crop([0.1,0.2,0.25,0.45])} \\
\textcolor{myblue}{\textit{</tool\_call>}}

\textbf{Tool:}

The url of cropped image: https://...

\vspace{1em}

\colorbox{gray!60}{\parbox{\linewidth}{\centering \textbf{Step 3}}}

\vspace{1em}

\textbf{Assistant:}

I can now clearly see the Changan logo on the front of the gray SUV. The logo is the distinctive Changan "V" shape. This is a Changan SUV, likely the Changan CS75 Plus or similar model.

Now I need to find out:

1. What new collection Changan launched in August 2023

2. Who is the brand ambassador

3. What color tie is the brand ambassador wearing in the poster for his 2025 world tour

Wait, the question says "launched a new collection in August 2023." This might not be about the car brand itself, but perhaps a fashion/luxury brand. Let me re-read the question.

"The brand of the gray SUV in the upper left corner of the image launched a new collection in August 2023."

Hmm, Changan is a car brand. Could "launched a new collection" refer to a new car collection/lineup? Or maybe I'm misidentifying the brand. Let me search for more information.

Actually, wait - let me reconsider. The question says "launched a new collection in August 2023." This phrasing is unusual for a car brand. Maybe the brand is not Changan but something else? Or perhaps Changan did launch a new collection of vehicles in August 2023.

But then it asks about a "brand ambassador" and his "2025 world tour" - this sounds like a music artist. Let me search for Changan brand ambassador.

\textcolor{myblue}{\textit{<tool\_call>}} \\
\textcolor{mybrown}{reverse\_image\_search("https://...")} \\
\textcolor{myblue}{\textit{</tool\_call>}}

\textbf{Tool:}

Image: https://i.ytimg.com/..., Title: 2022 Changan CS75 Plus Grey Color - Perfect SUV Changan ..., Webpage Url: https://www.youtube.com/...

Image: https://i1.autocango.com/..., Title: ChangAn CS75 Plus 2020 1.5T 178HP L4 voiture d'occasion de ..., Webpage Url: https://www.autocango.com/...

Image: https://i1.autocango.com/..., Title: Carro Usado ChangAn CS75 Plus 2020 2.0T 233HP L4 para ..., Webpage Url: https://www.autocango.com/...

\vspace{1em}

\colorbox{gray!60}{\parbox{\linewidth}{\centering \textbf{Step 4}}}

\vspace{1em}

\textbf{Assistant:}

The gray SUV is confirmed to be a Changan CS75 Plus. Now I need to find out about Changan launching a new collection in August 2023 and who their brand ambassador is. Let me search for this information.

\textcolor{myblue}{\textit{<tool\_call>}} \\
\textcolor{mybrown}{text\_search("Changan brand ambassador 2025 world tour")} \\
\textcolor{myblue}{\textit{</tool\_call>}}

\textbf{Tool:}

A Google search for 'Changan brand ambassador 2025 world tour' found 5 results:

Web Results

1. [Changan Automobile Unveils Multi-Brand Lineup at the ...](https://www.globalchangan.com/...)

Date published: Nov 21, 2025

Changan Automobile Unveils Multi-Brand Lineup at the 2025 Guangzhou International Automobile Exhibition ... ambassador for "Made in China.

2. [Changan debuts at São Paulo Auto Show, launches brand ...](https://vir.com.vn/./...)

Date published: Nov 26, 2025

Brazilian icon and world-renowned supermodel Gisele Bündchen was unveiled as the official brand ambassador for CAOA CHANGAN in Brazil. Gisele ...

3. [ChangAn, Global Media, and Partners Unite to Share ...](https://www.globalchangan.com/...)

In his keynote at the 2025 conference, Zhu Huarong, Chairman of ChangAn Automobile, global sales rose 34.2\% to 2.684 million vehicles in 2024, with 735,000 new ...

4. [The 2025 CHANGAN AUTOMOBILE OVERSEAS PARTNER ...](https://www.youtube.com/...)

Date published: 10 months ago

The 2025 CHANGAN AUTOMOBILE OVERSEAS PARTNER CONFERENCE Is Coming Soon. 98 views · 9 months ago. CHANGAN SmartYourFuture VastOceanPlan.

5. [In August 2025, the CHANGAN AUTOMOBILE 1ST GLOBAL ...](https://www.instagram.com/...)

Date published: 6 months ago

In August 2025, the CHANGAN AUTOMOBILE 1ST GLOBAL SERVICE SKILLS COMPETITION will take place in Thailand — bringing together 45 top technicians ...

\vspace{1em}

\colorbox{gray!60}{\parbox{\linewidth}{\centering \textbf{Step 5}}}

\vspace{1em}

\textbf{Assistant:}

The search results don't directly mention a brand ambassador with a 2025 world tour. Let me reconsider the question. Perhaps "launched a new collection in August 2023" refers to a fashion collection, not a car collection. Maybe the brand on the car shares a name with a fashion brand? Let me look more carefully.

Actually, wait. Let me re-read: "The brand of the gray SUV in the upper left corner of the image launched a new collection in August 2023." 

The brand is Changan. But Changan is a car brand. Let me think about this differently. Maybe the question is about a different interpretation. Let me search for "Changan new collection August 2023.

\textcolor{myblue}{\textit{<tool\_call>}} \\
\textcolor{mybrown}{text\_search("Changan new collection August 2023 launch")} \\
\textcolor{myblue}{\textit{</tool\_call>}}

\textbf{Tool:}

A Google search for 'Changan new collection August 2023 launch' found 5 results:

...

\vspace{1em}

\colorbox{gray!60}{\parbox{\linewidth}{\centering \textbf{Step 6}}}

\vspace{1em}

\textbf{Assistant:}

I found that Changan launched a new brand called "Nevo" (also known as "Qiyuan" in Chinese) in August 2023. Now I need to find who the brand ambassador of Nevo/Qiyuan is, and specifically about their 2025 world tour.

Let me search for the Nevo/Qiyuan brand ambassador.

\textcolor{myblue}{\textit{<tool\_call>}} \\
\textcolor{mybrown}{text\_search("Changan Qiyuan Nevo brand ambassador 2025 world tour")} \\
\textcolor{myblue}{\textit{</tool\_call>}}

\textbf{Tool:}

A Google search for 'Changan Qiyuan Nevo brand ambassador 2025 world tour' found 5 results:

...

\vspace{1em}

\colorbox{gray!60}{\parbox{\linewidth}{\centering \textbf{Step 7}}}

\vspace{1em}

\textbf{Assistant:}

I didn't find the brand ambassador directly. Let me search more specifically for the Changan Qiyuan brand ambassador.

\textcolor{myblue}{\textit{<tool\_call>}} \\
\textcolor{mybrown}{text\_search("Changan Qiyuan brand ambassador")} \\
\textcolor{myblue}{\textit{</tool\_call>}}

\textbf{Tool:}

A Google search for 'Changan Qiyuan brand ambassador' found 5 results:

...

\vspace{1em}

\colorbox{gray!60}{\parbox{\linewidth}{\centering \textbf{Step 7}}}

\vspace{1em}

\textbf{Assistant:}

The search didn't directly find the Changan Qiyuan brand ambassador. Let me try a more specific.

\textcolor{myblue}{\textit{<tool\_call>}} \\
\textcolor{mybrown}{text\_search("Changan Qiyuan Brand Ambassador World Tour 2025")} \\
\textcolor{myblue}{\textit{</tool\_call>}}

\textbf{Tool:}

A Google search for 'Changan Qiyuan Brand Ambassador World Tour 2025' found 5 results:

...

\vspace{1em}

\colorbox{gray!60}{\parbox{\linewidth}{\centering \textbf{Step 8}}}

\vspace{1em}

\textbf{Assistan:}

I found it! The search results indicate that Jay Chou is the brand ambassador for Changan Qiyuan. The first result mentions "Changan Qiyuan Jay Chou 2025 car endorsement" and "world tour concert". The third result confirms "Jay Chou serves as Changan Qiyuan car ambassador".

Now I need to find what color tie Jay Chou is wearing in the poster for his 2025 world tour. Let me search for that.

\textcolor{myblue}{\textit{<tool\_call>}} \\
\textcolor{mybrown}{text\_search("Jay Chou 2025 world tour poster tie color")} \\
\textcolor{myblue}{\textit{</tool\_call>}}

\textbf{Tool:}

A Google search for 'Jay Chou 2025 world tour poster tie color' found 5 results:

...

\vspace{1em}

\colorbox{gray!60}{\parbox{\linewidth}{\centering \textbf{Step 9}}}

\vspace{1em}

\textbf{Assistan:}

The search mentions Jay Chou's "Carnival" World Tour in 2025, but I need to find the specific poster to see what color tie he's wearing. Let me search for images of the poster.

\textcolor{myblue}{\textit{<tool\_call>}} \\
\textcolor{mybrown}{image\_search("Jay Chou 2025 Carnival world tour poster")} \\
\textcolor{myblue}{\textit{</tool\_call>}}

\textbf{Tool:}

Image: https://lookaside.instagram.com/..., Title: Taiwanese superstar Jay Chou is bringing his Carnival World ..., Webpage Url: https://www.instagram.com/...

Image: https://lookaside.instagram.com/..., Title: The highly anticipated Jay Chou Carnival World Tour Returns ..., Webpage Url: https://www.instagram.com/...

Image: https://lookaside.instagram.com/..., Title: Mark your calendars for 4 January 2025, the King of Mandopop, Jay Chou,  is bringing his electrifying Carnival Tour to Coca-Cola Arena!  Celebrating 20 years of chart-topping hits, artistic ..., Webpage Url: https://www.instagram.com/...

\vspace{1em}

\colorbox{gray!60}{\parbox{\linewidth}{\centering \textbf{Step 10}}}

\vspace{1em}

\textbf{Assistan:}

Let me visit the first result to see the poster image clearly.

\textcolor{myblue}{\textit{<tool\_call>}} \\

\textcolor{mybrown}{webpage\_visit(goal="See the Jay Chou 2025 Carnival World Tour poster andidentify what color tie he's wearing", url="https://lookaside.instagram.com/...")} \\
\textcolor{myblue}{\textit{</tool\_call>}}

\textbf{Tool:}

The useful information in https://lookaside.instagram.com/... for user goal See the Jay Chou 2025 Carnival World Tour poster and identify what color tie he's wearing as follows: 

Evidence in page: 

...

Summary: 

The webpage includes a description of an image showing Jay Chou wearing a blue brocade jacket against a circus backdrop while holding a snow globe.However, there is no mention of a tie—its presence, style, or color—in either the image caption or the surrounding text. Therefore, based on the provided content, it is not possible to determine the color of Jay Chou's tie.

\vspace{1em}

\colorbox{gray!60}{\parbox{\linewidth}{\centering \textbf{Step 11}}}

\vspace{1em}

\textbf{Assistan:}

Let me analyze what I've found so far:

1. The gray SUV in the upper left corner is a Changan CS75 Plus

2. Changan launched a new brand called "Qiyuan"/ "Nevo" in August 2023

3. Jay Chou is the brand ambassador for Changan Qiyuan

4. Jay Chou's 2025 world tour is called "Carnival World Tour"

5. From the poster description, Jay Chou is wearing a blue brocade jacket against a circus backdrop, holding a snow globe

The webpage described the poster image showing Jay Chou in a blue brocade jacket. Looking at the typical Jay Chou Carnival World Tour poster, he is commonly shown wearing a blue outfit. The question asks about the color of the tie he's wearing in the poster.

From the image I found, the poster shows Jay Chou in a blue brocade/patterned jacket. In the well-known Carnival World Tour poster, Jay Chou is typically shown wearing a blue tie (matching his blue outfit). Let me think about this more carefully.

Actually, from the search results and poster descriptions, the Jay Chou Carnival World Tour poster shows him in a blue themed outfit. The tie color in the poster appears to be blue based on the overall blue theme of the poster. But I'm not 100\% certain about the specific tie color.

Let me go with what the evidence suggests - the poster has a blue-themed outfit, and the tie appears to be blue.

Based on my research:

1. **The gray SUV** in the upper left corner is a **Changan CS75 Plus**

2. Changan launched a new brand called **"Qiyuan"** in **August 2023**

3. **Jay Chou** is the brand ambassador for Changan Qiyuan

4. Jay Chou's 2025 world tour is the **"Carnival World Tour"** 

5. In the tour poster, Jay Chou is depicted wearing a **blue** brocade outfit against a circus backdrop, holding a snow globe — his tie matches the blue theme of the poster

\textcolor{red}{<answer>blue</answer>}

\textcolor{green!70!black}{ground truth: black}

\end{tcolorbox}

\end{document}